\pdfoutput=1

\documentclass[11pt]{article}

\usepackage[preprint]{acl}

\usepackage{times}
\usepackage{latexsym}
\usepackage{hyperref}
\usepackage{amsmath}
\usepackage{algorithm}
\usepackage{algpseudocode}
\usepackage[T1]{fontenc}

\usepackage[utf8]{inputenc}

\usepackage{microtype}

\usepackage{inconsolata}

\usepackage{graphicx}
\usepackage{amsmath}

\title{PatentEdits: Framing Patent Novelty as Textual Entailment}

\author{
\textbf{Ryan Lee\textsuperscript{1}} \hspace{3em}
\textbf{Alexander Spangher\textsuperscript{1}} \hspace{3em}
\textbf{Xuezhe Ma\textsuperscript{1}} 
\\[0.5em] 
\textsuperscript{1}University of Southern California
\\[0.5em] 
\small{\textbf{Correspondence:} \href{mailto:ryantlee@usc.edu}{ryantlee@usc.edu}}
}

\begin{document}
\maketitle
\begin{abstract}
A patent must be deemed novel and non-obvious in order to be granted by the US Patent Office (USPTO).  If it is not, a US patent examiner will cite the prior work, or prior art, that invalidates the novelty and issue a non-final rejection. Predicting what claims of the invention should change given the prior art is an essential and crucial step in securing invention rights, yet has not been studied before as a learnable task. In this work we introduce the PatentEdits dataset, which contains 105K examples of successful revisions that overcome objections to novelty. We design algorithms to label edits sentence by sentence, then establish how well these edits can be predicted with large language models (LLMs). We demonstrate that evaluating textual entailment between cited references and draft sentences is especially effective in predicting which inventive claims remained unchanged or are novel in relation to prior art.


\end{abstract}

\section{Introduction}
\begin{figure}[t]
  \includegraphics[width=\columnwidth]{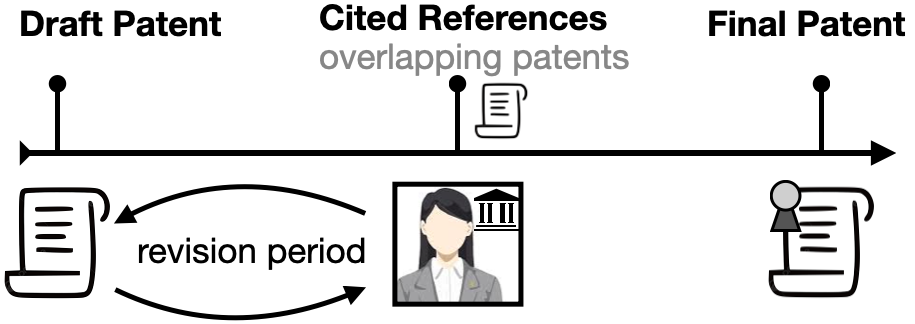}
  \caption{Simplified patent application timeline. PatentEdits aligns text data from the draft, cited references and final patents to understand patent revision.}
  \label{fig:timeline}
\end{figure}

Prior work in document edit prediction, such as in news domains \citep{spangher-etal-2022-newsedits}, do not consider how edits of a given document are influenced by the surrounding body of work. For inventors and examiners, considering the relation of a draft patent to prior work is a critical part of the patent application process, as shown in Fig. \ref{fig:timeline}. 

Predicting patent edits is also an important task: patents are critical protections of intellectual property that grant inventors the exclusive rights to make, use, and sell a disclosed invention for 20 years \cite{uspto2024utility}.  Unlike prior patent datasets such as the Harvard USPTO Dataset \citep{NEURIPS2023_b4b02a09} that focus on the patentability of the first application, we focus anticipating what claims need to be edited in order to be original with respect to the cited references, or the \textit{prior art}. Understanding how prior art influences patent edits would be of use to a majority of those applying for patents and the US patent examiners themselves: in a 2015 Yale Law study, \citealt{carley2015probability} found that 86\% of all patent applications are initially rejected by the Patent Office and then revised. 

To address the need to study how patents are successfully rewritten after the first application, we introduce PatentEdits, a dataset built to characterize patent revisions and the impact of cited inventive overlap. Our contributions are the following:
\begin{enumerate}
    \item We provide a corpus of 105k patents where text data from the draft, cited references, and final patent are aligned. We further algorithmically determine which patent sentences are \textit{Kept}, \textit{Edited}, or \textit{Deleted}, and ground these labels with human evaluation.
    \item We provide a procedure for adapting semantic retrievers for the patent domain with the PatentEdits dataset, by using an LLM aligned cited sentence as the positive example, the draft sentence as the anchor, the final sentence as the negative example, then fine-tuning retrievers with triplet loss.
    \item With our edit prediction experiments, we demonstrate that entailment-based approaches are an effective means of evaluating patent claim novelty. We report that classification improves by including the cited references and focusing on the entailment between cited and draft patent claims.
\end{enumerate}





\section{The PatentEdits Dataset}
\label{sct:dataset}
PatentEdits consists of 105,000 utility patents from 2007 to 2014.  Unlike prior patent datasets, each example in PatentEdits aligns the draft to the final granted patent text as well as the complete text of the patents or publications cited by the USPTO examiner. These cited references contain descriptions and ideas which invalidate the novelty of the draft patent, and by including them, we are able to use this context for edit prediction. An instance of the PatentEdits dataset is provided in Table \ref{tab:PatentEdits_instance}.

In this work, we focus specifically on the \textit{claim sentences} of the draft, cited, and final patents. This critical section of a patent describes the legal coverage of the invention claimed and is a primary focus during official review by the USPTO \cite{duening2020protecting}.

\subsection{Dataset Sources}
There exists no single bulk source that aligns both the draft and final claims as well as the cited references, so we extracted and aligned from 4 separate, publicly available USPTO datasets. Aligning the draft text to the final text is challenging as the draft patents are published separately from the final patents with different unique identifiers: the draft patent text only has a unique publication number, whereas the final granted patent only has a unique patent number. 

\subsection{Extracting Draft and Final Patent Text}
\label{sct:text-extraction}
First, to align the final patent text to the correct corresponding draft we download the Patent Examination Research Dataset \citep{Graham2015}, which we found provides a mapping between the publication number and the patent number.  We use SQL queries (Google BigQuery) based on publication number to extract the draft patent text, then queries based on patent number to extract the final patent text. We specifically choose to query from two USPTO's Patent Claims Research Datasets available on Google Cloud \citep{marco2016patent}, which includes pre-processed, cleaned patent sentences. We conduct a merge of the text data found from draft and from final patents, keeping only the examples with both. Because these cleaned and pre-processed text sources span from 2007-2014, so too does PatentEdits.

\begin{table}[t]
\centering
\begin{tabular}{p{0.3cm} p{6cm}}
\hline
\textbf{ID} & US Patent 7490780 \\
\hline
\textbf{D} & [ ``1. A radiation image projection apparatus comprising: an image acquisition unit\ldots configured to project the radiation image as visible light onto the object\ldots'',  \\
              &  ``2. The radiation image projection apparatus according to claim 1, further comprising: a radiation generating unit\ldots'']\\
\hline
\textbf{C} &[ ``1. An X-ray device \ldots wherein the indicator means includes an acoustic alarm device \ldots'\\
              & 2. An X-ray device \ldots wherein the indicator means is arranged on one of: the X-ray detector and the X-ray source.\ldots'']\\ \hline
\textbf{F} & [ ``1. A radiation image projection apparatus comprising: an image acquisition unit\dots and a mirror configured to transmit radiation and reflect visible light;\ldots'',  \\
              &  ``2. The radiation image projection apparatus according to claim 1, further comprising: a radiation generating unit\ldots'']\\ \hline
$\boldsymbol{\varepsilon}$ & ['edit', 'edit', 'del', 'keep', 'keep', 'keep', 'keep', 'keep', 'keep', 'edit', 'edit'] \\
\textbf{E} & [(0, 0), (1, 1), (3, 2), (4, 3), (5, 4), (6, 5), (7, 6), (8, 7), (9, 8), (10, 9)] \\
\hline
\end{tabular}
\caption{An instance in the PatentEdits dataset, which consists of the patent number \textbf{ID}, list of draft sentences \textbf{D}, a list of final patent sentences \textbf{F}, a list of cited reference sentences \textbf{C}, a list of edit labels for each draft sentence $\boldsymbol{\varepsilon}$, and \textbf{E}, a list of found sentence edges.}

\label{tab:PatentEdits_instance}
\end{table}

\subsection{Extracting Examiner Cited References}
To obtain the text data from examiner cited references, we download the USPTO Office Action Citations Dataset and filter for the 35 U.S.C. 102 and 35 U.S.C. 103 objections, which correspond to office rejections for novelty and non-obviousness \cite{hellmann2024rejections}. This dataset only contains a list of unique identifiers, so as we did in Section \ref{sct:text-extraction}. we query for the text data using those identifiers.

To make attribution of inventive overlap easier, we filtered the dataset for patents that have 1 or 2 cited references (48\% of the original examples) and only include examples where the full citation context is found. We plan to release the full PatentEdits dataset and scripts in our project repository. 

\subsection{Edit Label Extraction}
\label{subsec:extraction}
\begin{figure}[t]
  \includegraphics[width=\columnwidth]{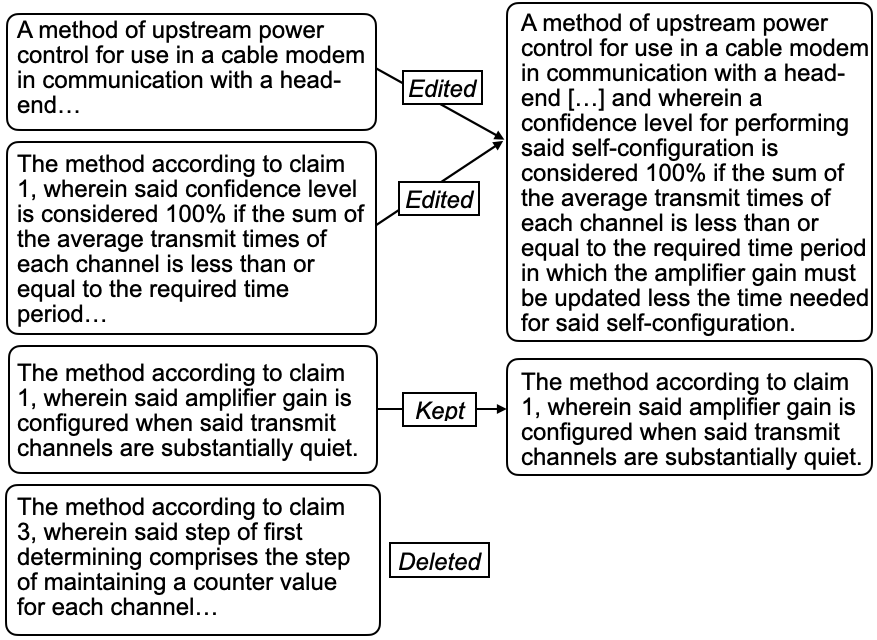}
  \caption{Shown are the extracted edit labels for US Patent 8677435. On the left are draft claims and on the right are final patent claims, with edges denoting a sentence match.}
  \label{fig:edit_track}
\end{figure}

In order to study trends in patent revisions, we apply algorithms to synthetically label which sentences of the patent are \textit{Kept}, \textit{Edited}, or \textit{Deleted}. We construct a bipartite graph and find edges between sentences of the draft and final and determine the \textit{edit actions} possible on draft sentences. As shown in Fig. \ref{fig:edit_track}, after the sentence matches are found,  edit actions are determined by the following set of rules: 
\begin{itemize}
    \item \underline{\textit{Kept}}: a draft sentence is labeled as \textit{Kept} if it is matched to a final patent sentence and its similarity score is above a \textit{Kept} threshold. We validate this threshold further with an additional human study.
    \item \underline{\textit{Edited}}: a draft sentence is labeled as \textit{Edited} if matched to a final patent sentence and its similarity score is below the \textit{Kept} threshold.
    \item \underline{\textit{Deleted}}: a draft sentence is labeled as \textit{Deleted} if there is no final sentence with sufficient similarity above a \textit{Deleted} threshold.
\end{itemize}

To find the best edges, we explore greedy algorithms to find these edges based on similarity scores between draft and final sentences. In this work, we do not consider sentence splits: we limit the number of outgoing edges to only one, from each draft sentence. In all approaches we have a \textit{Deleted} threshold on similarity score, where if no edge is found with a sufficient similarity, that draft sentence is considered \textit{Deleted}.

We first evaluate using the same sentence matching following \citep{spangher-etal-2022-newsedits} using BLEU-4 \cite{papineni2002bleu} as the scoring method. We consider this approach to be ``draft-side greedy'' as it matches every draft sentence to the grant sentence it has the highest similarity score with. In addition to BLEU-4 as the scoring method, we explore using BLEU-1, METEOR \citep{banerjee-lavie-2005-meteor}, CHRF \citep{popovic-2015-chrf}, and BERTscore \citep{zhang2019bertscore}. 

In our next set of experiments, we explore a ``final-side greedy'' approach, assigning a final patent sentence to the draft sentence it has the highest similarity score with. Similarly to the draft-side greedy approach, we explore using Rouge-L, Rouge-1, METEOR, CHRF, and BERTscores as the similarity score, and select the best score based on alignment with human annotations. 

Motivated by the observation that many draft sentences can be \textit{Edited} into a single final sentence, we implement a match-and-cover algorithm for final-side greedy approaches to ensure that all relevant draft sentences are properly attributed to a given final patent sentence.  More specifically, we exhaustively search for all draft sentences that could match with the remaining grant sentence, after a longest common subsequence (LCS) is removed following a match. We detail the exact logic in Alg. \ref{alg:match-cover}.

\begin{algorithm}
\caption{Match-and-Cover Procedure for Final-Side Greedy Matching}
\begin{algorithmic}[1]
\State \textbf{Input:} Set of draft sents $D$, set of final sents $F$, matching threshold $\tau$, fraction limit $\epsilon$
\State \textbf{Output:} Matches of draft to final sentences

\For{each final sentence $f \in F$}
    \While{text in $f$ left > than frac limit $\epsilon$}
        \State Find best unmatched draft $d^*$ for $f$
        \State Identify LCS between $d^*$ and $f$
        \If{match score is >= to threshold $\tau$}
            \State Remove LCS from $f$
            \State Attribute $d^*$ to $f$

            \State Add $d^*$ to ``matched'' set
        \Else
            \State \textbf{break}
        \EndIf
    \EndWhile
\EndFor
\end{algorithmic}
\label{alg:match-cover}
\end{algorithm}

\begin{table*}[t]
\centering
\begin{tabular}{l ll ll ll}
\hline
    & \multicolumn{2}{c}{\textbf{Embedding}} & \multicolumn{2}{c}{\textbf{Lexical}} & \multicolumn{2}{c}{\textbf{Ngram}} \\
 Algorithm         & Scoring Method & F1 & Scoring Method & F1 & Scoring Method & F1 \\ \hline
Draft-side Greedy & Roberta-Large      & 83.0  & ChrF    & 60.7  & BLEU-1  & 80.0  \\
                           & Deberta-large-mnli & 82.7  & METEOR  & 82.0  & BLEU-4  & 82.9  \\ \cline{1-7}
Final-side Greedy & Roberta-large      & 87.8  & ChrF    & 35.6  & Rouge-1 & 87.1  \\
                           & Deberta-large-mnli & 88.6  & METEOR  & 87.8  & Rouge-L & \textbf{90.3} \\ \hline
\end{tabular}
\caption{
Sentence Matching F1 scores against human annotations using embedding, lexical, and ngram-based scoring methods and greedy matching.}
\label{tab:sentence-matching}
\end{table*}

\subsection{Human Evaluation of Edit Labels}
\label{subsec:sentence-matching}
To evaluate the quality of the automatic matching between draft and final sentences of the document, we instruct human annotators to match sentences between draft and final patent documents for 90 patents in the PatentEdits dataset. 

For all the sentences of the draft and final patent, annotators are instructed to match a given draft sentence to a granted sentence if they have substantial overlap in meaning, even in instances where the inventive detail has been paraphrased in the revised granted sentence. If there is no substantial conceptual overlap between the two sentences, or if it is unclear how the two sentences are semantically related, we instruct the annotator not to match the sentences. We use our sentence-matching algorithms on the same 90 human annotated patents, then report the F1 scores with the human annotated labels as the ground truth label. 

An edge between a draft and final sentence can mean a sentence has been either \textit{Edited} or \textit{Kept}. To determine the difference with an automatic means (after sentence edges are found), we set up an additional human study: we select a random subset of 100 examples of matched sentence pairs and evaluate if they are \textit{Kept} or \textit{Edited}. We then asked annotators to mark the paired draft and final sentence as \textit{Edited} if there was substantial conceptual difference between the two sentences, or \textit{Kept} if the two sentences were exactly the same, with the exception of different claim numbering or typos. We then use this human evaluation study to determine the \textit{Kept} threshold for all our different approaches. 

\subsection{Automatic Edit Label Performance}
We report the resulting performance of these algorithms and scoring methods in Table \ref{tab:sentence-matching}. We find our best sentence scoring algorithm is a final-side greedy approach which uses Rouge-L scoring, with a \textit{Deleted} threshold of 0.45 and a remaining fraction threshold of 0.3. Our best sentence matching algorithm achieves an F1 score of 90.3 against human annotations. This F1 score is comparable to the best matching F1 score of 89.5 seen in NewsEdits \cite{spangher-etal-2022-newsedits} with a BERT-based algorithm and between revisions of news articles.

We found that using BLEU-4 with a \textit{Kept} threshold of 0.88 resulted in the best agreement with the human labeled \textit{Edited} vs. \textit{Kept} examples, with an F1 score of 0.91 (\textit{Edited} F1) and 0.97 (\textit{Kept} F1). To obtain the edit labels for the entire dataset, we apply the final-side greedy, Rouge-L matching algorithm and the \textit{Kept} threshold (BLEU-4 of 0.88) to the entire PatentEdits dataset.

\section{Training Semantic Retrievers}
During patent prosecution, the examiner and patent writer may directly discuss the specific claims which must be changed, as well as the specific overlap in the prior patent cited; Thus these exchanges are not readily available to researchers, as these interviews can be conducted over the phone and are not fully on public record. However, in PatentEdits, because we have the text data of all the cited patents in the patent application process, it is possible to extract that specific context. To extract the specific context from the cited references, we rank the sentences from the cited documents based on their semantic similarity to a given draft claim and extract only the most relevant ones.

\begin{figure}[t]
  \includegraphics[width=\columnwidth]{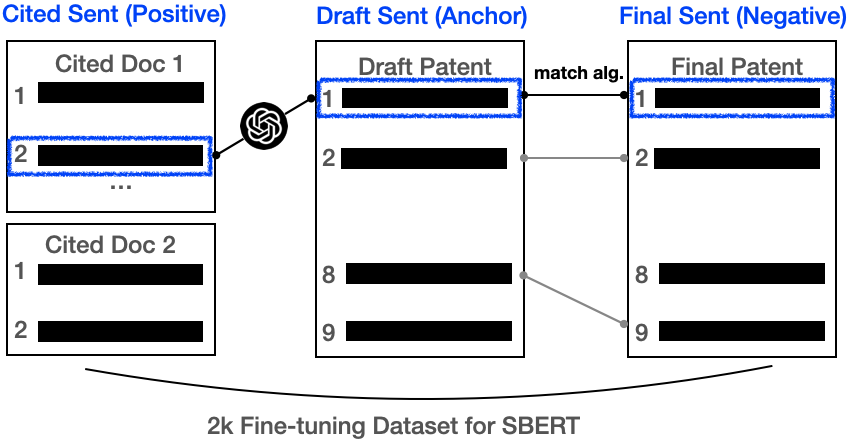}
  \caption{We obtain high quality retrievals of relevant citations with an LLM, treat it as the positive, the draft sentence as the anchor, and the final sentence as the negative, for triplet loss. We treat the final patent sentence as a negative example as we assume it has been rewritten to be semantically different from the draft. One training example for the retriever is one triplet of sentences.}
  \label{fig:triplet_loss_dataset}
\end{figure}

We are motivated to use retrievers to rank, of all cited sentences, which are most similar to a given draft sentence, then extract the top-k candidates. However, without specific fine-tuning retrievers may not find the correct passages or find irrelevant ones.  In this work, we fine-tune sentence-BERT \citep{reimers-2019-sentence-bert} to semantically retrieve relevant inventive detail, by using labels extracted by GPT4o and from PatentEdits. We outline the following general procedure for adapting off-the-shelf retrievers to the patent domain below and in Fig \ref{fig:triplet_loss_dataset}:
\begin{enumerate}
    \item First, we extract from our training split (see Section \ref{sec:edit_predictions}) a subset of 2000 \textit{Edited} sentences.
    \item Next, for each draft sentence in the subset, we prompt an LLM (in our case GPT4o) to copy the cited sentence the draft claim is most likely to be infringing on.
    \item With the 2K dataset, we use the extracted cited, draft, and final claim as the positive, anchor, and negative examples to train a semantic encoder with triplet loss \citep{schroff2015facenet}. 
    \item We then use the lightweight, fine-tuned sentenceBERT (SBERT) to retrieve the top cited sentence from the cited documents for the rest of the dataset.
\end{enumerate}

We use triplet loss to specifically leverage an important aspect of the dataset: by sentence matching and edit labeling, we have found the final sentences have been edited to semantically distance themselves from their draft versions. The triplet loss is defined as:
\begin{equation*}
\mathcal{L}(A, P, N) = \max(0, d(A, P) - d(A, N) + \alpha)
\label{eq:triplet_loss}
\end{equation*}
By using triplet loss, we train our encoder to map draft sentences (A) closer to the relevant cited references (P), while distancing from the semantically changed final patent sentences (N). We use this fine-tuned retriever to extract the most relevant cited reference sentence for every draft sentence in our dataset, and set up entailment experiments for edit prediction.

\section{Patent Edit Classifiers}
\label{sec:edit_predictions}
In this section we establish baselines for classifying the future edit of a given patent sentence given the cited references and the text of the draft sentence. For these experiments, we define 10k random subset of the full dataset, with an 80-10-10 train-validation-test split. We explore small and large models, vary the context size, as well as the model objective.

We ensure that the 10k subset covers the same time span as the full 105k dataset, and that the experimental splits have the same relative ratio of edit classes. We filter out for patents that have completely been rewritten and patents that were not revised at all. The experiments in this section are intended to illustrate how the included cited references and edit labels can be leveraged, and suggest the importance of the cited reference to the edit prediction task. 

\subsection{Edit Classification Experiments}
\paragraph{Sentence-only Edit Prediction} Given only the context of the draft sentence, we seek to predict the most likely the edit for a given draft sentence. Here, we investigate whether it is possible to predict the edit action given only the claim sentence, such as by identifying vague conceptual language that could not possibly be novel.

\paragraph{Sentence+Citation Edit Prediction}
Given both the context of the draft sentence and the most semantically similar cited reference sentences, we predict the edit action on the draft sentence. This experiment explores the signal that examiner-cited references have in influencing the edit outcome. To incorporate the most relevant cited sentence, we naively concatenate it to the end of the draft sentence, and include token delimiters. Specifically we add ``[claim]'' before the draft sentence and ``[cited]" before the cited reference sentence.

\paragraph{Window Edit Prediction} So far, we have only considered the context from a single sentence and/or a single cited reference sentence. In this experiment, we explore how increasing the context window and allowing models to process more context can improve edit label predictions. To prepare this context for an LLM, we simply concatenate the previous, current and following draft sentences into a text chunk and use as label the edit of the center draft sentence. As a starting point, we use a window of three draft sentences to predict one edit.

\begin{table*}[t]
\centering
\begin{tabular}{llllllll}
\hline
\textbf{Context} & \textbf{Model} & \textbf{Weighted }&\textbf{Micro} &\textbf{Macro} &\textbf{Kept} &\textbf{Edited} &\textbf{Deleted} \\
\hline
Sent & Roberta-Base & 50.3 &49.6 & 46.1 & 60.0 & 41.3 & 37.0 \\
Sent+Cit & Roberta-Base & 50.4 &49.7 & 46.4 & 59.7 & 41.7 & 37.4 \\
Window & Roberta-Base & 52.3 & 51.9 & 47.7 & 62.8 & 42.7 & 37.8 \\ \hline
Sent & Roberta-Large & 54.4 &54.4 & 49.8 & 65.4 & 45.4 & 38.7 \\
Sent+Cit & Roberta-Large & 54.0 &53.3 & 49.7 & 64.0 & 45.4 & 37.6 \\ \hline
(Cit,Sent) & Roberta-Large-MNLI & 55.5 & 57.3 & 49.1 & \textbf{70.1} & 42.7 & 34.5 \\
(Cit,Sent) & BART-Large-MNLI & \textbf{55.6} & \textbf{57.4} & 49.5 & 69.8 & 42.7 & 35.9 \\
\hline
\end{tabular}
\caption{
Test F1 scores for edit prediction on the experimental 10K dataset. We see that simply concatenating the closest extracted citation to the draft text does not improve results, but treating the citation and draft sentence as an entailment pair (\textit{premise}, \textit{hypothesis}), using models pre-trained on entailment, and mapping predicted entailment classes to edit classes greatly improves overall F1 score, especially for the \textit{Kept} category.
}
\label{edit-prediction-results}
\end{table*}

\paragraph{Cited to Draft Entailment}
In this experimental set-up, we re-frame the edit prediction task as classifying textual entailment. If the draft claim \textit{Entails} from the most relevant cited claim then it infringes and should be \textit{Edited} or \textit{Deleted}. If the draft claim \textit{Contradicts} the cited claim premise, than the draft claim can be \textit{Kept} as it is original or novel. 

To validate the mapping between entailment labels and edit labels, we first conducted a Chi-squared test of independence between the predicted entailment labels and the edit labels using an entailment model without fine-tuning (microsoft/deberta-large-mnli) \citep{he2021deberta}. The test yielded a p-value 2.6e-22, which is significantly lower than our significance level of $\alpha = 0.05$. As a result, we reject the null hypothesis that the predicted entailment labels and the edit labels are independent and conclude that there is a statistically significant association between the two sets of labels.

For the entailment experiments we fine-tune Hugging Face \citep{wolf-etal-2020-transformers} transformer models Roberta-Large-MNLI and BART-Large-MNLI, which are pre-trained on the Multi-Genre Natural Language Inference (MultiNLI) dataset \citep{williams2018broad}. During training, we consider the most relevant cited sentence as the premise and the draft sentence as the hypothesis. To adapt the labels to edit prediction, we simply map the \textit{Contradiction} label to \textit{Kept}, \textit{Neutral} to \textit{Edited}, and \textit{Entailment} to \textit{Deleted}.

\subsection{Additional Experimental Set-up}
\label{subsec:clf-setup}
For these sentence-level prediction tasks, we primarily used the RoBERTa architecture \citep{liu2019roberta}, an optimized BERT-based transformer. For all of the three context scenarios, we use under-sampling of the majority class on the training dataset to mitigate class imbalance in the dataset. For experiments with Roberta-base models we use a same batch size of 32, 6 training epochs, and a learning rate of 2e-5. For the Roberta-large models, we reduce the batch size to 16 to fit the training on a 40GB GPU, reduce the number of training epochs to 3, but keep the same learning rate of 2e-5. We used a learning rate warm-up of 500 steps. 

\section{Edit Prediction Results}
As shown in Table \ref{edit-prediction-results}, simply concatenating citations is not enough to improve draft edit prediction. Across the small and large models, we don't observe a significant change in performance by simply including the cited reference text as additional input. 

Our most notable result is that entailment between the citation and draft sentence significantly improves overall edit prediction specifically for the \textit{Kept} class, which corresponds to the \textit{Contradicts} entailment label. Using the entailment between the cited and draft sentence gives us the strongest weighted F1 score of all the models, however we do note that for the \textit{Deleted} class, the entailment models perform worse than the others.

We also see the expected trends: larger models outperform smaller ones, and larger context windows improve performance. More specifically, we see that Roberta-Large performs better than Roberta-Base for both with and without citation context, and we also see that the windowed context performs better than any other of the Roberta-Base model contexts. 

The F1 scores we see are comparable to what is reported in other research works on document edit prediction, such as NewsEdits \citep{spangher-etal-2022-newsedits} and LASERTAGGER \citep{malmi-etal-2019-encode}.

\section{Discussion}
In the classification experiments, including the cited references improved predictions on which sentences would be edited when we framed edit prediction as an entailment evaluation: we interpret this as confirming our assumption that the cited references provide critical information on how the patent should be rewritten. In these experiments, we use lightweight models and relatively small contexts, only the context of a single draft sentence and at most 2 additional sentences. As our windowed experiment suggests, including more of the draft document and cited references could further improve the classification performance. 

Our results suggest that in general, the \textit{Edited} and \textit{Deleted} classes are harder to predict, as they have lower classification F1 scores for all experiments than \textit{Kept}. We do not believe this is simply due to \textit{Kept} being the majority class, as we account for class imbalance by under-sampling the majority class and equalizing the number of each edit class examples in training. Instead, we hypothesize that once the set of offending claim sentences are identified (the ones not \textit{Kept}), the decision about which of the remaining to edit or delete may either be arbitrary or dependent on other factors beyond inventive overlap with the cited references.

\section{Related Work}
Pre-existing patent datasets for machine learning such as the Harvard USPTO Patent Dataset \citep{NEURIPS2023_b4b02a09} focus on classifying initial patentability, or predicting patent field class. In contrast, we construct PatentEdits to understand how the prior cited references influence the revision of patent claims.

The definition and extraction of sentence-level edit labels extends upon the work of \citet{spangher-etal-2022-newsedits} in the News domain: we adapt these methodologies for the patent domain by focusing on using examiner cited references to predict edits and focusing on predicting the revised patent claims text. 

\section{Conclusion}
In this work we curate a dataset of 105k patents to characterize document level change as a function of prior work, aligning text data from the draft, cited references, and final versions. We enable edit prediction by algorithmically labeling patent sentences (or claims) into edit categories such as \textit{Kept}, \textit{Edited}, or \textit{Deleted} and ground our labels with extensive human evaluation. We also provide a procedure for using PatentEdits to adapt semantic retrievers to the patent domain. By leveraging an LLM-aligned cited sentence as the positive example, the draft sentence as the anchor, and the final sentence as the negative example, we fine-tune the retrievers through triplet loss. Our edit prediction experiments validate the effectiveness of entailment-based approaches in assessing patent claim novelty. We demonstrate that incorporating cited references and focusing on entailment between cited and draft claims significantly improves classification performance.




\section*{Ethical Considerations}
\subsection*{Limitations }
The edit actions in PatentEdits are determined based on rules and automatic metrics and verified with human evaluation. While the annotators were able to manually verify truthfulness for a subset of examples, the quality and correctness of the automatic means may further improve with expert evaluation, i.e. by patent agents or USPTO examiners.

In this work we do not consider predicting the ``added" claims, but they do appear and can be tracked with PatentEdits. We consider \textit{Added} claims as substantively differen from classifying draft claim outcome, however we note that other works such as NewsEdits or LASERTAGGER do define tasks related to anticipating whether a sentence will be added before or after a given sentence.

With our preliminary studies on sentence-only context, we do not utilize the unique independent-dependent claim relationships: claim independence refers to the aspect that some sentences in a patent will refer and extend off of other sentences, i.e. ``wherein the golf glove of claim 1 further comprises of a velcro fastener."

\subsection*{Privacy and Risks}
We do not believe there to be any significant privacy risks associated with this dataset as patents are a matter of public record, and PatentEdits is aggregated from bulk datasets shared by the USPTO for the express purpose of research into the patent prosecution process. Although the USPTO Office Action dataset does contain personal identifiers for patent agents and examiners, only the examiner cited references were collected from that data source.

\subsection*{Computational Resources and Libraries}
The PatentEdits dataset was processed with a TPU from Google Colab with 334GB of memory as well as with Google BigQuery. We share the processing code to obtain the PatentEdits dataset from the original sources, however extracting from scratch will require those resources. The fine-tuning experiments in this work are conducted using a NVIDIA V100 GPU with 40GB of GPU memory. The use of GPT4 requires OpenAI credits, and a total of \$25 was expended for experiments and predictions with prompting.

We use HuggingFace libraries and models in this work, such as RoBERTa for edit prediction and encoders sentence-BERT from the Transformers library for extracting most similar edit examples as well as cited references. For evaluation, we utilize publicly available NLP libraries such as NLTK, scikit-learn, bert-score and rouge.

\bibliography{custom}

\end{document}